\documentclass[runningheads]{llncs}
\usepackage[T1]{fontenc}
\usepackage{amsmath,amssymb,amsfonts}
\usepackage{stmaryrd}
\usepackage{algorithmic}
\usepackage{multirow}
\usepackage{utfsym}
\usepackage{subcaption}
\usepackage{booktabs}
\usepackage{array}
\usepackage{bm}
\usepackage{graphicx}
\usepackage[colorlinks,
            linkcolor=blue,
            anchorcolor=blue,
            citecolor=blue]{hyperref}
\usepackage{hyperref}
\begin{document}
\title{Dynamic Pseudo Label Optimization in Point-Supervised Nuclei Segmentation}
\titlerunning{DoNuSeg: Dynamic Pseudo Label Optimization}
%
\author{Ziyue Wang\inst{1, 2} \thanks{Ziyue Wang, Ye Zhang and Yifeng Wang contributed equally.}  \and
Ye Zhang\inst{1} $^\star$ \and
Yifeng Wang\inst{3} $^\star$ \and Linghan Cai\inst{1} \and Yongbing Zhang\inst{1} \thanks{Corresponding author}\\
\email 200111326@stu.hit.edu.cn, zhangye94@stu.hit.edu.cn, wangyifeng@stu.hit.edu.cn, 
ceilinghans@gmail.com, 	ybzhang08@hit.edu.cn}
\authorrunning{Ziyue Wang, Ye Zhang, Yifeng Wang, Linhan Cai and Yongbing Zhang}

%
\institute{School of Computer Science and Technology, Harbin Institute of Technology (Shenzhen), 
\and  Department of Electrical and Computer Engineering, National University of Singapore,
\and School of Science, Harbin Institute of Technology (Shenzhen).
}

%

%
\maketitle             
\begin{abstract}

Deep learning has achieved impressive results in nuclei segmentation, but the massive requirement for pixel-wise labels remains a significant challenge. To alleviate the annotation burden, existing methods generate pseudo masks for model training using point labels. However, the generated masks are inevitably different from the ground truth, and these dissimilarities are not handled reasonably during the network training, resulting in the subpar performance of the segmentation model.
To tackle this issue, we propose a framework named DoNuSeg, enabling \textbf{D}ynamic pseudo label \textbf{O}ptimization in point-supervised \textbf{Nu}clei \textbf{Seg}mentation. 
Specifically, DoNuSeg takes advantage of class activation maps (CAMs) to adaptively capture regions with semantics similar to annotated points.
To leverage semantic diversity in the hierarchical feature levels, we design a dynamic selection module to choose the optimal one among CAMs from different encoder blocks as pseudo masks.
Meanwhile, a CAM-guided contrastive module is proposed to further enhance the accuracy of pseudo masks. 
In addition to exploiting the semantic information provided by CAMs, we consider location priors inherent to point labels, developing a task-decoupled structure for effectively differentiating nuclei.
Extensive experiments demonstrate that DoNuSeg outperforms state-of-the-art point-supervised methods. The code is available at https://github.com/shinning0821/MICCAI24-DoNuSeg.
\end{abstract}

\section{Introduction}
Nuclei instance segmentation in whole-slide images (WSIs) is crucial for uncovering tumor microenvironment and thus informing relevant decisions in disease treatment \cite{lu2018nuclear,natarajan2020segmentation,zhang2024seine,hou2019robust}. Recently, deep learning techniques \cite{ronneberger2015u,mahmood2019deep,stringer2021cellpose,feng2021mutual,huang2023affine,choudhuri2023histopathological} have promoted nuclei segmentation. However, the success of the segmentation algorithms is contingent on the availability of high-quality imaging data with corresponding pixel-wise labels provided by experts. The annotating process is time-consuming and labor-intensive, limiting the development of models. Meanwhile, point labels that annotate nuclei with single points effectively reduce the annotation cost, making it essential to develop point-supervised segmentation methods.

\begin{figure*}
    \vspace{-0.5cm}
	\centering
	\includegraphics[width=0.95\textwidth]{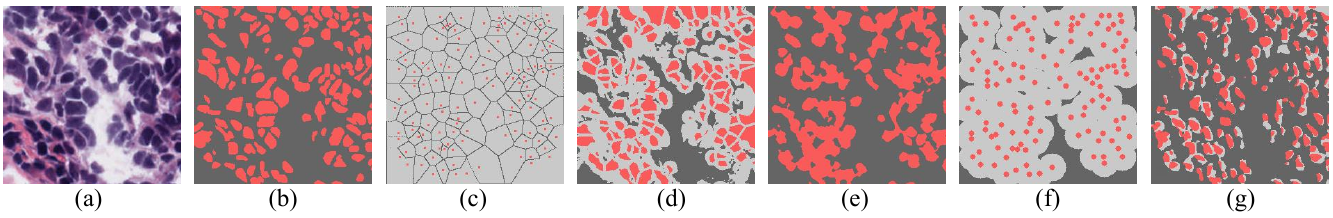} 
	\caption{\small (a) image input; (b) ground truth; (c) Voronoi label; (d) cluster label; (e) LSM label; (f) our initial label $M$; (g) our optimized label $P$. In (b)-(g), red, dark gray and light gray pixels denote nuclei, background and ignored areas, respectively.} 
     \vspace{-0.5cm}
	\label{label}
\end{figure*}

Existing point-supervised methods \cite{scnet,zhao2020weakly} generally adopt a two-stage framework, first utilizing the biological morphology of nuclei to generate pixel-wise pseudo masks, then training the segmentation model. For precise nuclei segmentation, current research investigates various approaches to improve the quality of the pseudo masks. As shown in Figure \ref{label}(c)(d), \cite{qu2020weakly,tian2020weakly} integrates the Voronoi diagram for mask generation, which considers the distance between points to distinguish overlapping instances and then generates cluster labels in separate regions. \cite{zhang2022ddtnet} develops a level-set method (LSM) to further consider the nuclei's topology as shown in Figure \ref{label}(e). However, these algorithms inevitably bring noisy labels due to the enormous variation in nuclei shape, color, and distribution. Meanwhile, these methods lack effective solutions for handling inaccurate labels. This oversight may impair the model training, leading to insufficient nucleus feature representation. To this end, devising reliable pseudo label enhancement in the training phase is critical for point-supervised nuclei segmentation.


Out of the advantages of targeting class-related areas, class activation maps (CAMs) are widely adopted for weakly supervised segmentation methods under natural scenes \cite{selvaraju2017grad,zhou2016learning}. During training, CAMs are gradually optimized and can increasingly localize foreground regions. Therefore, we argue that CAMs have the potential to serve as pseudo-labels. However, the direct application of CAMs for labels encounters great challenges. Firstly, nuclei are densely distributed in pathological images, while CAMs tend to capture the most salient regions, resulting in frequently missed detection. Secondly, nuclei present low contrast with the surrounding tissue, posing difficulty in determining instance boundaries using low-resolution CAMs. Therefore, it is imperative to enhance the quality of CAM generation and address the limitations of CAMs in distinguishing instances.



Motivated by the above discussions, this paper presents a \textbf{D}ynamic pseudo label \textbf{O}ptimization method in point-supervised \textbf{Nu}clei \textbf{Seg}mentation (DoNuSeg). DoNuSeg takes advantage of the location priors provided by point labels and the semantic-level representation of CAMs, decoupling nuclei instance segmentation into object detection and semantic segmentation. To alleviate the miss-detection problem, we develop a novel Dynamic CAM Selection (DCS) module that incorporates the hierarchical features of the encoder for CAM generation, enriching the nucleus-related features and obtaining more activated foreground areas. To suppress the CAMs' uncertainty, DoNuSeg adopts a CAM-guided contrastive learning (CCL) module highlighting the representation differences between nuclei and the surrounding tissues, thereby accurately distinguishing nuclei boundaries. Overall, our contributions can be summarized as following aspects:
\begin{itemize}
\item[$\bullet$] We propose a novel weakly supervised nuclei instance segmentation framework termed DoNuSeg, which effectively leverages CAMs to achieve dynamic optimization of the pseudo label.
\item[$\bullet$] We develop a pseudo label optimizing method that measures the accuracy of CAMs generated by different feature levels and adaptively selects the optimal CAM for label generation. 
\item[$\bullet$] We integrate a contrastive learning approach that utilizes the location information provided by points to widen the gap between nuclei and tissues, refining the feature representation and improving CAMs' location accuracy.
\item[$\bullet$] Extensive experiments demonstrate the superiority of our method, outperforming state-of-the-art methods on three public datasets. 
\end{itemize}

\section{Related Work}
\subsection{Supervised Nuclei Segmentation}
In recent years, deep learning has benefited many areas, particularly image segmentation \cite{cheng2022masked,xie2021segformer,yang2024exploring,zhang2022understanding,lai2024lisa,yang2023improved}. Convolutional neural networks (CNNs) have also been widely applied to pathology images \cite{ronneberger2015u,zhou2018unet++,aubreville2024domain,lu2024h2aseg}.
Due to the severe overlapping of nuclei and the high similarity between foreground and background regions in pathological images, the performance of these methods often drops when applied to nuclei segmentation. To address the issue, DCAN try to distinguish different instances by predicting the contour of the nuclei \cite{chen2016dcan}.  Hover-Net \cite{Graham2019} adds additional branches to predict the horizontal and vertical gradient maps, which helps distinguishing different nuclei. Cellpose \cite{stringer2021cellpose} predicts a binary map and the gradients of topological maps that indicate if a given pixel is inside or outside of regions of interest to refine the cell shape. CDNet \cite{he2021cdnet} and SEINE \cite{zhang2024seine} also constructs a direction map to represent the spatial relationship between pixels within the nuclei. Sams-net \cite{graham2018sams} and Triple-Unet \cite{zhao2020triple} are proposed to use H-channel images as input to predict better boundary. However, these methods rely on pixel-wise annotated masks to generate additional supervisory information and thus cannot be applied to scenarios when using point labels. 

\subsection{Weakly-Supervised Segmentation}
In many real-world scenarios, detailed annotated data is limited \cite{yang2024unified,lai2021semi,tian2020prior,}, especially in pathological images.
Weakly supervised methods conducted in pathological images mainly use box annotation, scribble annotation, and point annotation. 
Point labels are the most extensive in actual scenes. However, it is hard to train the network solely with annotated points. Therefore, existing methods generate pixel-wise pseudo labels based on the nuclei morphology. Voronoi is widely used to generate pseudo masks for segmentation \cite{dong2020towards,zhao2020weakly,guo2021learning,tian2020weakly,chamanzar2020weakly,yoo2019pseudoedgenet}. Besides, \cite{nishimura2021weakly} utilizes propagating detection map to segment fluorescence images, while \cite{zhang2024boundary,zhang2024dawn,wang2023semi}  propose different strategies to improve the quality of pseudo labels. However, these methods often perform semantic segmentation without distinguishing different instances, and the quality of the generated pseudo labels greatly influences the performance. In H\&E stained histopathology images, the morphology and density of nuclei vary a lot in different datasets. Thus the pseudo labels often contain much noise when applied to different situations and degrade the model's performance.

\section{Methods}
\label{sec:method}

\begin{figure*}
    \vspace{-0.5cm}
	\centering
	\includegraphics[width=0.95\textwidth]{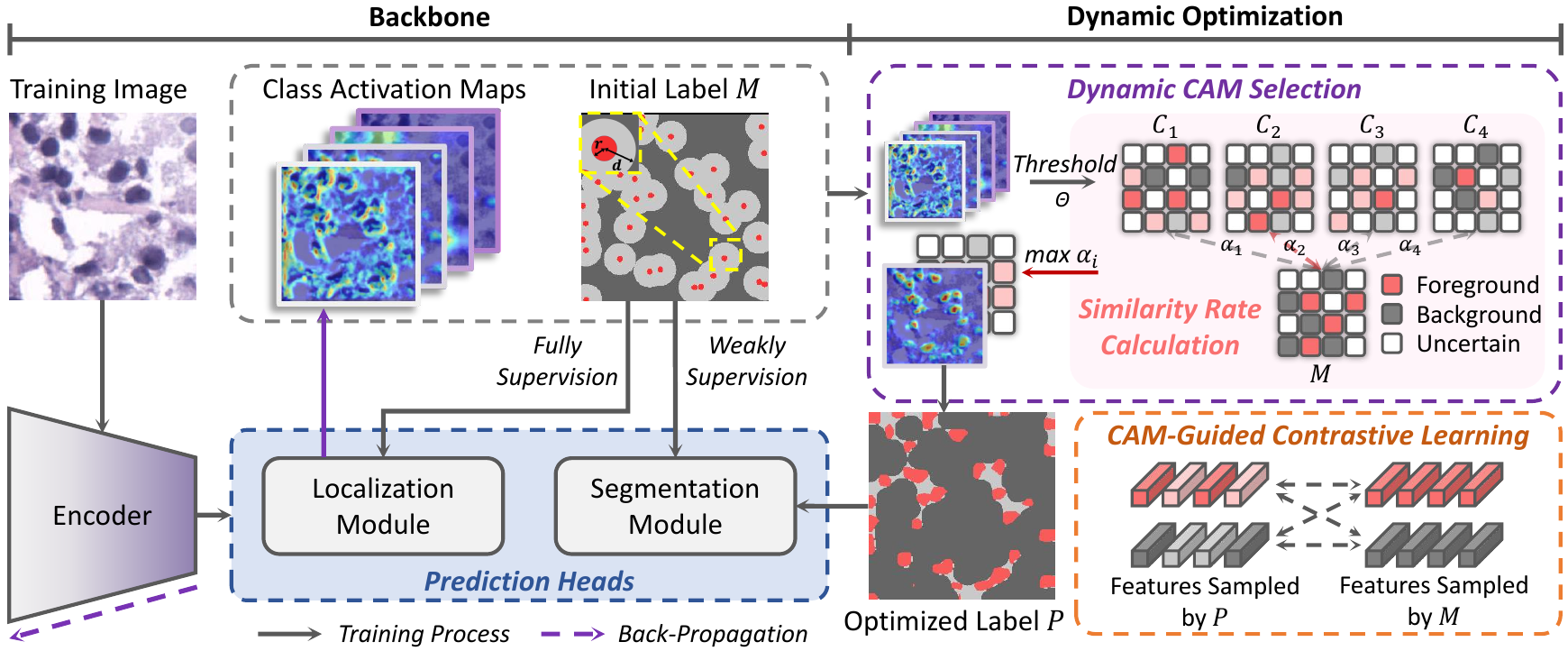} 
	\caption{\small Overview of our DoNuSeg method, which utilizes a Dynamic CAM Selection (DCS) module and a CAM-guided Contrastive Learning (CCL) module to dynamically select and optimize pseudo labels.} 
    \vspace{-0.5cm}
	\label{stru}
\end{figure*}
\begin{figure*}[ht]
	\centering
	\includegraphics[width=0.95\textwidth]{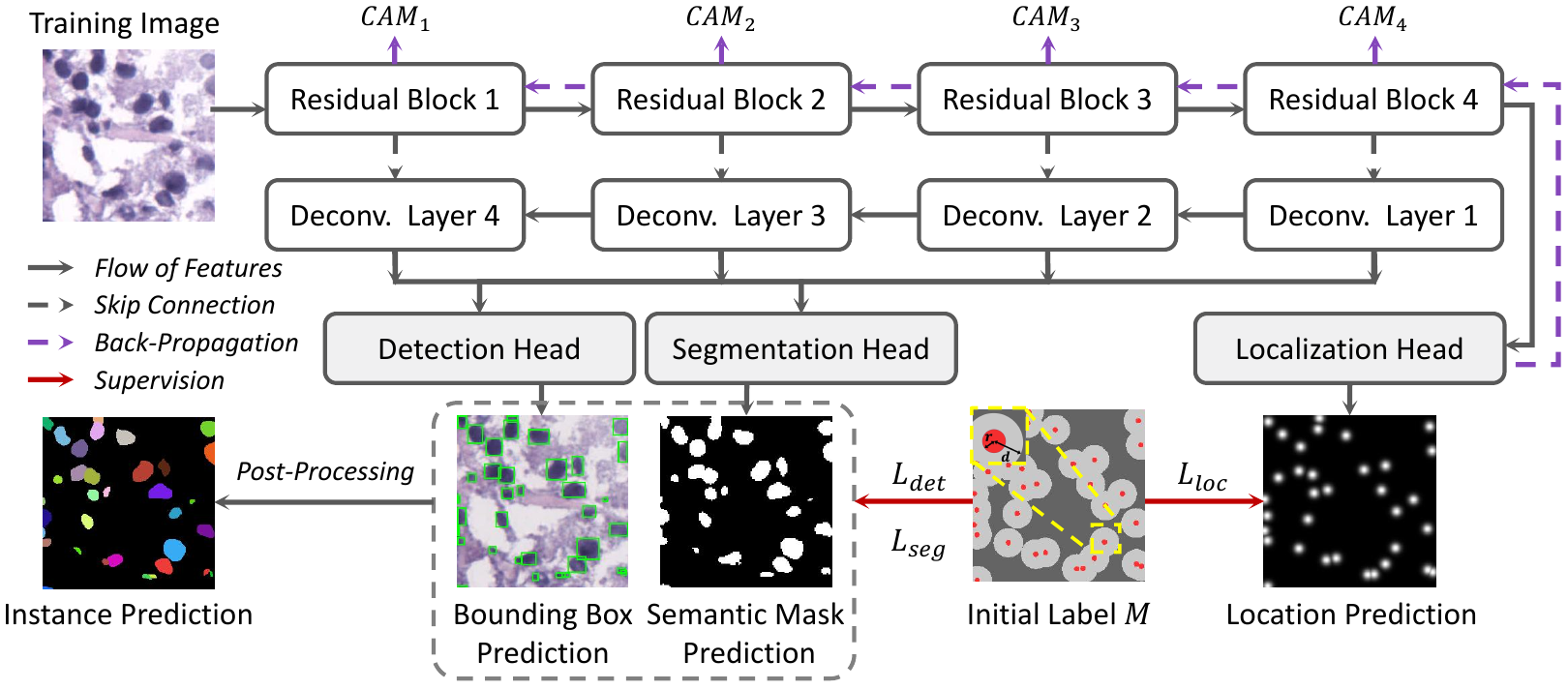} 
	\caption{\small Backbone structure of DoNuSeg. The detection and segmentation head takes hierarchical feature levels in the decoder as their input.} 
    \vspace{-0.5cm}
	\label{backbone}
\end{figure*}

Our DoNuSeg develops a dynamic pseudo label optimization method to solve the challenges in point-supervised nuclei segmentation by the proposed DCS and CCL module as shown in Figure \ref{stru}. To further utilize the location prior, we also take a task-decoupled structure as shown in Figure \ref{backbone}, which combines the detection task and semantic segmentation task to achieve instance segmentation.



\subsection{Backbone}

Point annotations are challenging for pixel-wise segmentation but can be utilized to train fully supervised agent tasks to generate CAMs. CAMs offer valuable insights into the model's focus on crucial foreground regions, providing valuable guidance for training segmentation networks. However, CAMs only capture semantic information and require additional assistance differentiating individual instances. To address this limitation, we propose a decoupled instance segmentation method, as illustrated in Figure \ref{backbone}. Our approach leverages the positional priors obtained from point annotations to accurately predict bounding boxes, facilitating the distinction of instances within the semantic masks.



We take FPN \cite{lin2017feature} as the backbone with a ResNet50 \cite{he2016deep} encoder. The decoder has a shared detection and segmentation head for each feature level. For the detection head, our design is based on the efficient detector FCOS \cite{tian2019fcos} while the segmentation head is composed of four convolutional layers. The predicted bounding boxes are used to distinguish instances from the semantic masks. 

We first calculate the pseudo bounding box for point labels following dense object detection in natural scenarios \cite{wang2021self}, which is used to compute the loss of detection heads $L_{det}$ following FCOS \cite{tian2019fcos}  (details seen in the supplementary materials).
As shown in Figure \ref{label}(f), we obtain the initial label $M$ for segmentation by assuming that pixels within $r$ units around the point labels are foreground and pixels more than $d$ units away from the point labels are background. Other pixels are ignored while training to avoid introducing noise. The segmentation loss is computed as follows:
\begin{equation}
    \mathcal{L}_{seg} = -\frac{1}{|\Omega_M|}\sum_{i \in \Omega_M} [(1-M_i)log(1-Y_i) + M_ilog(Y_i)],
    \label{seg_loss}
\end{equation}
where $Y_i$ and $M_i$ denote segmentation prediction $Y$ and initial label $M$ at the $i$th pixel, and $\Omega_M$ is the set of non-ignored pixels in $M$. For the generation of CAMs, a localization head following the encoder is employed to conduct fully supervised point localization, which consists of three fully connected layers and is trained by an MSE loss $\mathcal{L}_{loc}$.
The total loss function is computed as follows:
\begin{equation}
    \mathcal{L} = \mathcal{L}_{det} + \mathcal{L}_{seg} + \mathcal{L}_{loc} + \omega_1 \mathcal{L}_{dcs} + \omega_2 \mathcal{L}_{ccl},
    \label{total_loss}
\end{equation}
where $\mathcal{L}_{dcs}$ and $\mathcal{L}_{ccl}$ are the loss of the proposed DCS and CCL module and will be introduced in the following subsections. $\omega_1$ and $\omega_2$ are hyperparameters. 


\subsection{Dynamic CAM Selection Module}
As shown in Figure \ref{stru}, CAMs can reflect the attention area of the encoder and exhibit different semantic information depending on specific layers. Previous studies \cite{selvaraju2017grad,zhou2016learning} primarily utilize CAMs generated by the encoder's last layer, which has limited coverage of the foreground regions and coarse-grained boundaries resulting from upsampling from the small-size deep feature map.
However, the encoder's intermediate layers' CAMs, which capture more nuclei and can provide fine-grained information, are often ignored. Therefore, the DCS module is utilized to choose the proper CAM dynamically.
As shown in Figure \ref{stru}, we first filter the generated CAM by a threshold $\theta$ to get a binary map $C$:
\begin{equation}
C_i = \left\{
  \begin{array}{ll}
    1, &  CAM_i > \theta, \\
    0, & CAM_i \leq 1-\theta,
    \\
    \text{ignored}, & 1-\theta < CAM_i \leq \theta,
  \end{array}
\right.
\end{equation}
where $CAM_i$ and $C_i$ denote CAM and $C$ at the $i$th pixel. The similarity rate $\alpha$ of $C$ is defined as follows:
\begin{equation}
    \alpha = \frac{1}{|\Omega_M|} \sum_{i \in \Omega_M} M_iC_i+ \Bar{M_i}\Bar{C_i},
   \label{cam_err}
\end{equation}
where $M_i$ denote the initial label $M$ at the $i$th pixel, and $\Omega_M$ is the set of non-ignored pixels in $M$. We choose $C$ with the maximum $\alpha$ as the optimized pseudo label $P$ to dynamically supervise segmentation network training by:
\begin{equation}
    \mathcal{L}_{dcs} = -\frac{\alpha_P}{|\Omega_P|}\sum_{i \in \Omega_P} [(1-P_i)log(1-Y_i) + P_ilog(Y_i)],
    \label{cam_loss}
\end{equation}
where $P_i$ and $Y_i$ denotes $P$ and the segentation prediction $Y$ at the $i$th pixel, $\alpha_P$ is the similarity rate of $P$, and $\Omega_P$ is the set of non-ignored pixels in $P$.
\vspace{-0.2cm}
\subsection{CAM-guided Contrastive Learning Module}
To better segment nuclei boundaries, we propose a CAM-guided Contrastive Learning module to enhance intra-class coherence and inter-class discrimination of nuclei and background features. By aligning the CAM's attention regions with the initial label $M$,  CAM concentrates more on nuclei and less on the background, which enhances the accuracy of CAM. Details are described below:

We use a projector $g$ which consists of four $3\times3$ convolutional layers and a three-layer MLP to preserve critical contextual information following \cite{chen2020simple}. The features outputted by the encoder are enhanced by $g$ and then upsampled to the size of the original image as the enhanced feature map $Z$.
Let $Z_i$, $M_i$, $P_i$ denote $Z$, the initial Label $M$ and the optimized label $P$ at the $i$th pixel respectively, nuclei and background feature sets are defined as $\mathcal{F}^{+} = \{Z_i \in Z | M_i=1\}$ and $\mathcal{F}^{-} = \{Z_i \in Z | M_i=0\}$. 
The anchor features $a^+$ and $a^-$ are computed by the average of $\mathcal{F}^{+}$ and $\mathcal{F}^{-}$. Positive and negative feature sets can be sampled by $P$:  $\mathcal{S}^{+} = \{Z_i \in Z | P_i=1\}$ and $\mathcal{S}^{-} = \{Z_i \in Z | P_i=0\}$.
The pixel-wise contrastive learning loss is computed by:
\begin{equation}
    \mathcal{L}_{ccl} = \alpha_P(\mathcal{L}_{con}(a^+,S^+,S^-) + \mathcal{L}_{con}(a^-,S^-,S^+)),
    \label{con_loss_2}
\end{equation}
where $\alpha_P$ is the similarity rate of $P$. $\mathcal{L}_{con}$ is defined as follows:
\begin{align}
\mathcal{L}_{con}(q,U,V) = -\frac{1}{|U|} \sum_{u \in U} [\phi(q,u)/ \tau 
- \log (e^{\phi(q,u)/ \tau} + \sum_{v \in V} e^{ \phi(q,v)/ \tau })],
\label{con_loss_1}
\end{align}
where $\tau$ is a hyperparameter and $\phi$ denotes the consine similarity, $q$ is an anchor feature and $U,V$ are the similar and dissimilar feature sets. 

As described in Section 2.1, the initial pseudo label $M$ hardly contains noise, so anchor features can be regarded as the ground truth for nuclei and background features. The proposed method can make the features of the foreground and background pixels in $P$ closer to the corresponding anchor features, thus enhancing the feature representation. 


\section{Experiments and Results}
\vspace{-0.3cm}
\begin{table}[h]\small
\vspace{-0.5cm}
  \centering
  \renewcommand{\arraystretch}{1.1}
  \caption{Performance comparison (\%) with SOTA point-supervised methods. The best performance is shown in \textbf{bold}, and the second is \underline{underlined}.}
    \resizebox{0.91\textwidth}{!}{
    \begin{tabular}{l|l|c|c|c|c|c}
    \hline
    Datasets & Methods   &  DICE \quad    & AJI\quad  & DQ\quad   & SQ\quad    & PQ\quad   \\ \hline
    CryoNuSeg  & WeakSeg \cite{qu2020weakly}  & 59.08$\pm$2.27     & 29.07$\pm$1.81 & 34.97$\pm$1.15 & 62.91$\pm$1.42 & 23.78$\pm$1.51 \\
    & PseudoEdgeNet \cite{yoo2019pseudoedgenet}     & 60.42$\pm$1.71  & 36.78$\pm$2.17 & 35.03$\pm$3.32 & 63.10$\pm$1.53 & 22.37$\pm$1.29 \\
    & MaskGA-Net \cite{guo2021learning}        & 65.94$\pm$2.85 & \underline{40.13}$\pm$0.73 & 41.18$\pm$1.81 & \textbf{67.47}$\pm$0.89 & \underline{28.19}$\pm$1.13 \\
    & DDTNet \cite{zhang2022ddtnet}            & \textbf{68.32}$\pm$1.97        & 34.05$\pm$0.93 & \underline{41.83}$\pm$0.81 & \underline{66.86}$\pm$1.40 & 27.91$\pm$2.17 \\
    & SC-Net \cite{scnet}            & 63.17$\pm$0.74        & 38.63$\pm$1.28 & 38.82$\pm$0.44 & 65.32$\pm$1.34 & 25.43$\pm$2.03 \\
    & \textbf{DoNuSeg} (Ours) & \underline{67.22}$\pm$1.30   & \textbf{44.08}$\pm$0.77      &  \textbf{46.41}$\pm$1.67     &  65.74$\pm$1.31    &  \textbf{30.58}$\pm$1.53     \\
    \hline
    CoNSeP & WeakSeg \cite{qu2020weakly}           & \textbf{63.32}$\pm$1.16   & \underline{33.41}$\pm$1.34 & \underline{34.97}$\pm$2.10 & \underline{64.91}$\pm$1.60 & \underline{23.17}$\pm$0.58 \\
    & PseudoEdgeNet \cite{yoo2019pseudoedgenet}     & 33.07$\pm$2.24        & 22.07$\pm$1.10 & 14.38$\pm$1.82 & 52.54$\pm$0.67 & 15.26$\pm$1.11 \\
    & MaskGA-Net \cite{guo2021learning}        & 28.69$\pm$1.61            & 20.70$\pm$1.55 & 19.97$\pm$0.36 & 52.37$\pm$2.03 & 18.81$\pm$1.29 \\
    & DDTNet \cite{zhang2022ddtnet}            & 58.82$\pm$1.29        & 29.58$\pm$2.18 & 28.33$\pm$0.25 & \textbf{65.20}$\pm$1.77 & 20.67$\pm$0.43 \\
    & SC-Net \cite{scnet}        & 49.95$\pm$1.55            & 31.70$\pm$0.43 & 27.77$\pm$0.17 & 57.12$\pm$0.76 & 21.58$\pm$2.05 \\
    & \textbf{DoNuSeg} (Ours) & \underline{60.20}$\pm$2.10   & \textbf{36.39}$\pm$1.77      &  \textbf{36.68}$\pm$0.74      &  62.33$\pm$1.83      &  \textbf{23.95}$\pm$0.99 \\
    \hline
    TNBC & WeakSeg \cite{qu2020weakly}           & \underline{66.22}$\pm$0.33         & 44.82$\pm$1.91 & 50.45$\pm$0.37 & \underline{67.91}$\pm$1.58 & 33.05$\pm$2.34 \\
    & PseudoEdgeNet \cite{yoo2019pseudoedgenet}     & 48.27$\pm$0.40        & 32.83$\pm$1.38 & 30.31$\pm$2.74 & 54.76$\pm$0.72 & 20.26$\pm$1.92 \\
    & MaskGA-Net \cite{guo2021learning}        & 55.01$\pm$2.07        & 36.16$\pm$1.09 & 40.26$\pm$1.40 & 56.24$\pm$2.45 & 27.13$\pm$1.98 \\
    & DDTNet \cite{zhang2022ddtnet}            & \textbf{67.88}$\pm$1.59        & \underline{47.83}$\pm$2.01 & \underline{58.56}$\pm$0.57 & 66.79$\pm$2.26 & \underline{40.08}$\pm$0.76 \\
    & SC-Net \cite{scnet}        & 63.24$\pm$1.91        & 44.75$\pm$1.92 & 53.62$\pm$1.21 & 58.38$\pm$0.39 & 35.29$\pm$0.67 \\
    &\textbf{DoNuSeg} (Ours) & 63.99$\pm$1.18   & \textbf{50.06}$\pm$0.94      &  \textbf{58.77}$\pm$0.69      &  \textbf{68.75}$\pm$1.34     &  \textbf{40.64}$\pm$0.54 \\
    \hline
    \end{tabular}}
  \label{tab:cro_weak}
  \vspace{-0.5cm}
\end{table}

\subsection{Datasets and Metrics}
\vspace{-0.1cm}
We evaluate the proposed method on three public datasets, namely CryoNuSeg \cite{mahbod2021cryonuseg}, ConSeP \cite{Graham2019}, and TNBC \cite{naylor2018segmentation}. CryoNuSeg contains 30 images sampled from 10 organ tissues with the size of $512\times 512$. ConSeP includes 41 images sampled from colon patients with the size of $1000 \times 1000$. TNBC consists of 50 images from 11 breast cancer patients with a size of $512\times 512$. Datasets are divided into training, validation, and test sets in a ratio of 3:1:1. All images are then cropped into $256\times256$ sized patches with an overlapping of 128 pixels.

We adopt five widely used metrics for quantitative evaluation: DICE, Aggregated Jaccard Index (AJI) \cite{kumar2017dataset}, Detection Quality (DQ), Segmentation Quality (SQ), and Panoptic Quality (PQ) \cite{kirillov2019panoptic}. The higher value is better for these metrics. To avoid randomness, we adopt 5-fold cross-validation and report the average values and the standard deviation in the testing set.



\vspace{-0.2cm}
\subsection{Implementation Details}
\vspace{-0.2cm}
Our experiments are implemented on PyTorch 1.10.0 using an Nvidia RTX 3090 GPU. 
We adopt an SGD optimizer for model training with a learning rate of 0.01, a momentum of 0.9, and a weight decay of 0.0005. Each model is trained for up to 40 epochs with a mini-batch size of 8.
We set hyperparameters $r=4$, $d=20$, $\tau=1$, $\omega_1$= 0.5, $\omega_2$= 2, and $\theta=0.8$. Online data augmentation is employed to alleviate the over-fitting, including random flipping, random rotation, and random cropping.

\subsection{Comparisons with State-of-the-art Methods}
\vspace{-0.2cm}
We compare DoNuSeg with popular point-supervised nuclei segmentation methods: WeakSeg \cite{qu2020weakly}, PseudoEdgeNet \cite{yoo2019pseudoedgenet}, MaskGA-Net \cite{guo2021learning}, DDTNet \cite{zhang2022ddtnet}, and SC-Net \cite{scnet}. 
It is worth noting that MaskGA-Net and PseudoEdgeNet are designed for semantic segmentation, thus we obtain the instance mask by applying post-processing following \cite{Graham2019}. 

\begin{figure*}[h!]
	\centering
	\includegraphics[width=0.98\textwidth]{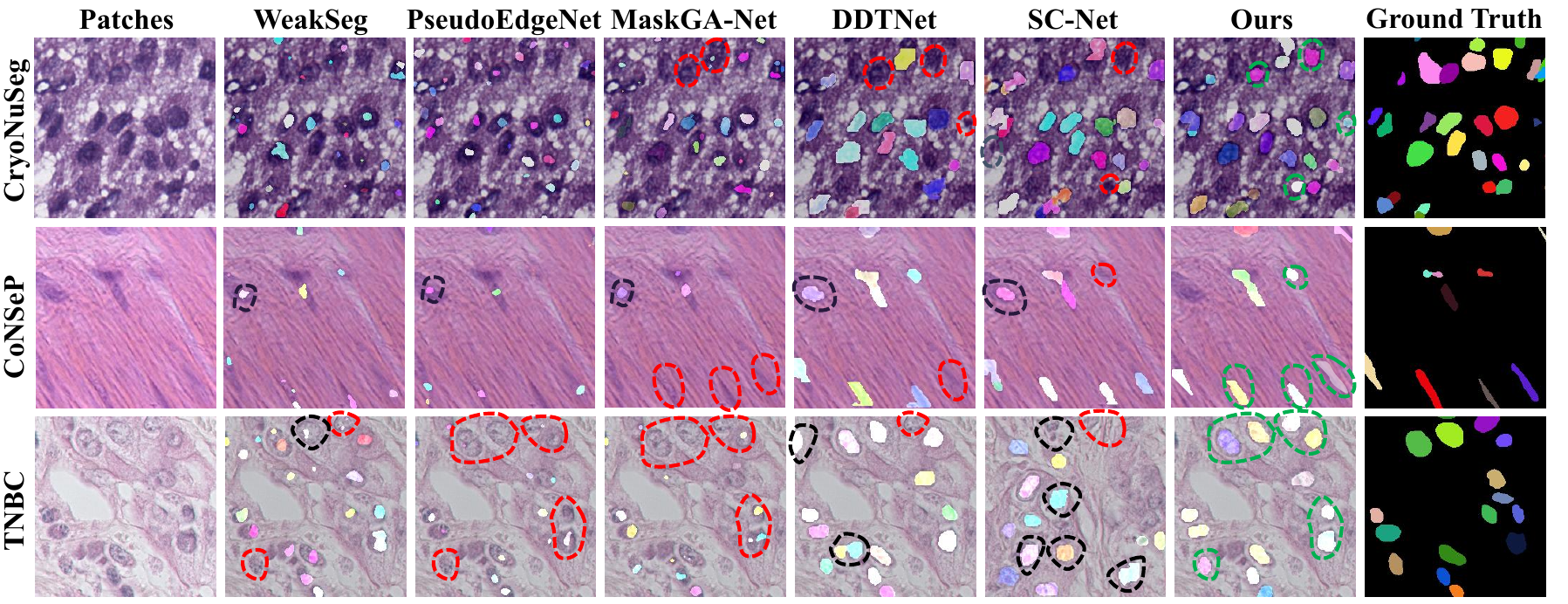} 
	\caption{\small Visualization comparison of segmentation results on three datasets. Red and black circles indicate the false negative (FN) and false positive (FP) errors. Green circles denote how DoNuSeg corrects these errors.} 
    \vspace{-0.8cm}
	\label{img_result}
\end{figure*}
\subsubsection{Quantitative Evaluation.} Table \ref{tab:cro_weak} presents performance comparisons in terms of five metrics. It can be seen that previous methods present poor performance due to the absence of correction for noisy pseudo labels.
In contrast, our method outperforms state-of-the-art methods on AJI, DQ, and PQ across all the datasets. Notably, DoNuSeg achieves significant improvements in terms of AJI, surpassing the second-best by 3.9\%, 2.9\%, and 2.2\% on the three datasets, respectively.


\vspace{-0.3cm}
\subsubsection{Qualitative Evaluation.} Figure \ref{img_result} displays the visual comparison results. As challenging datasets, CryoNuSeg and CoNSeP have a low distinction between nuclei and background tissue. Thus, the generated pseudo-labels based on morphology measure often involve much noise and lead to numerous FN and FP errors. Surprisingly, DoNuSeg can dynamically select and optimize pseudo labels, thus performing well on these challenging datasets. 

\begin{table*}[htbp]
  \centering
  \renewcommand{\arraystretch}{1.1}
  \caption{Effects (\%) of $\mathcal{L}_{dcs}$ and $\mathcal{L}_{ccl}$ on CryoNuSeg and CoNSeP.}
  \resizebox{0.9\textwidth}{!}{
    \begin{tabular}{cc|ccccc|ccccc}
    \hline
    \multirow{2}{*}{$\mathcal{L}_{dcs}$} & \multirow{2}{*}{$\mathcal{L}_{ccl}$} & \multicolumn{5}{c|}{CryoNuSeg}   & \multicolumn{5}{c}{CoNSeP}     \\ 
    \cline{3-12}
    &                                            & DICE & AJI & DQ & SQ & PQ & DICE & AJI & DQ & SQ & PQ \\ \hline
    \multicolumn{1}{c}{} &     \multicolumn{1}{c|}{}     & 60.93        & 41.78       & 42.68      & 63.93      & 27.32      & 53.69        & 35.51       & 35.77      & 55.18      & 22.85      \\
    \multicolumn{1}{c}{\checkmark} &     \multicolumn{1}{c|}{}      & \underline{66.48}        & \underline{43.82}       & \textbf{46.41}      & 64.99      & \underline{30.20}      & \underline{57.94}        & 35.75       & 35.92      & 60.14      & \underline{23.36}      \\
    \multicolumn{1}{c}{} &     \multicolumn{1}{c|}{\checkmark}       & 64.02        & 43.71       & 45.66      & \underline{65.60}      & 30.06      & 52.98        & \textbf{36.53}       & \underline{36.55}      & \underline{60.43}      & 23.12      \\
    \multicolumn{1}{c}{\checkmark} &     \multicolumn{1}{c|}{\checkmark}     & \textbf{67.22}        & \textbf{44.08}       & \textbf{46.41}      & \textbf{65.74}      & \textbf{30.58}      & \textbf{60.20}        & \underline{36.39}       & \textbf{36.68}      & \textbf{62.33}      & \textbf{23.95}   \\ \hline  
    \end{tabular}
    }
    \vspace{-0.3cm}
    \label{tab:ablation_cro}
\end{table*}
\subsection{Ablation Study}
We conduct ablation experiments on CryoNuSeg and CoNSeP to prove the effectiveness of the proposed method. As shown in Table \ref{tab:ablation_cro}, the method significantly improves when adding $\mathcal{L}_{dcs}$ and $\mathcal{L}_{ccl}$ and achieves the best performance when both are added. 
This shows that DCS and CCL improve the training performance of the segmentation network by improving the quality of the pseudo-label. 

\noindent
\begin{minipage}[c]{0.45\textwidth}
\centering
\tiny
\vspace{0cm}
\captionof{table}[tb]{\small Effects (\%) of the CAM selection strategy on CryoNuSeg.}
\resizebox{\linewidth}{32pt}{
\begin{tabular}{c|ccccc}
\hline
block & DICE   & AJI    & DQ     & SQ     & PQ     \\ \hline
$1$ & 61.95   & 40.14 & 39.88 & 64.81 & 27.14 \\
$2$ & 64.53   & 40.91 & 41.16 & 65.35 & \underline{29.06} \\
$3$ & \underline{67.10}   & 41.44 & 40.97 & \textbf{66.02} & 27.37 \\
$4$ & 62.84 & \underline{41.73} & \underline{45.60} & 64.38 & 28.35 \\ \hline
ours & \textbf{67.22} & \textbf{44.08} & \textbf{46.41} & \underline{65.74} & \textbf{30.58} \\
\hline
\end{tabular}}\label{tab:ablation_block}
\vspace{0.3cm}
\end{minipage}
\hspace{15pt}
\begin{minipage}[c]{0.45\textwidth}
\centering
\tiny
\vspace{0.4cm}
\captionof{figure}[tb]{\small Effects (\%) of different $r$ and $d$ when generating $M$ on CryoNuSeg.}
\includegraphics[width=1.05\textwidth]{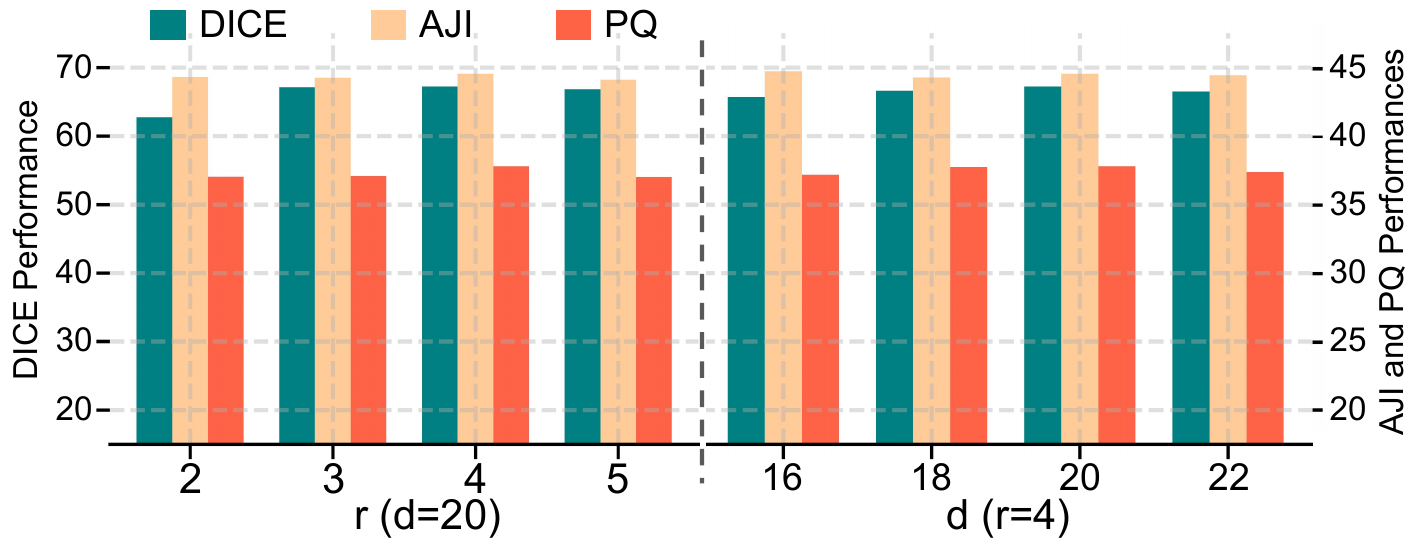}
\label{bar}
\vspace{0.3cm}
\end{minipage}

Table \ref{tab:ablation_block} shows the effect of the CAM selection strategy in the DCS module. It can be seen that the proposed method can combine semantic information in hierarchical features and achieve the best performance. Notably, compared to merely using  CAM generated by the fourth block as in previous methods, the performance improves by 4.4\%, 2.4\%, and 2.1\% on DICE, AJI, and PQ, respectively. Furthermore, as shown in Figure \ref{bar}, when changing the hyperparameters $r$ and $d$ in generating the initial label $M$, the model's performance is almost unaffected, which shows that our method is robust to the selection of parameters.
\section{Conclusion}
This paper proposes a point-supervised nuclei segmentation framework, DoNuSeg, to reduce the cost of pixel-level annotations. DoNuSeg utilizes CAMs to achieve a dynamic optimization mechanism of the noisy pseudo labels. A DCS module and a CCL module are proposed to dynamically select and optimize CAMs and gradually correct the pseudo label. To better distinguish nuclei, we develop a task-decouple structure to leverage location priors in point labels. Experiments show that our method achieves SOTA performance, and the ablation study shows the effectiveness of the proposed method. In conclusion, DoNuSeg provides fresh insights for point-supervised nuclei instance segmentation.

\subsubsection{Acknowledgements.} This work was supported in part by the National Natural Science Foundation of China under 62031023 \& 62331011; and in part by the Shenzhen Science and Technology Project
under GXWD20220818170353009, and in part by the Fundamental Research Funds for the
Central Universities under No.HIT.OCEF.2023050.


%
%
%
\bibliographystyle{splncs04}
\bibliography{miccai}

\end{document}